\documentclass[11pt, leqno]{article}
\usepackage[english]{babel}

\usepackage[utf8]{inputenc}
\usepackage[T1]{fontenc}
\usepackage{dsfont}
\usepackage{hyperref}       
\usepackage{url}            
\usepackage{booktabs}       
\usepackage{amsfonts}       
\usepackage{nicefrac}       
\usepackage{amsmath, amssymb}
\usepackage[a4paper,hscale=0.72,vscale=0.8,centering]{geometry}
\usepackage{setspace}
    
\usepackage{authblk}
\usepackage{amsfonts}
\usepackage{bbm}
\usepackage{dsfont}
\usepackage{graphicx}
\usepackage{float}
\usepackage{xcolor}
\usepackage{wrapfig}
\usepackage{tikz}
\usepackage{pst-node}
\usepackage{algorithm}
\usepackage{listings}
\usepackage{fancyvrb}
\usetikzlibrary{fit,positioning,bayesnet}
\usetikzlibrary{automata,arrows}
\usetikzlibrary{backgrounds}

\usepackage{subfigure}
\usepackage[noend]{algorithmic}

\usepackage{natbib}
\usepackage{hyperref}
\definecolor{bred}{rgb}{0.8,0,0}
\hypersetup{colorlinks,linkcolor={blue},citecolor={black},urlcolor={blue}}
\setcitestyle{square,numbers,comma}
\usepackage{cleveref}

\def\sZ{{\mathsf{Z}}}

\def \0{\mathbf 0}

\def \cL{\mathcal L}

\def \cU{\mathcal U}

\def\md{{\mathrm d}}
\def\KL{{\mathrm{KL}}}
\def\cZ{\mathcal{Z}}

\def\cY{\mathcal{Y}}

\def\b0{\mathbf 0}

\def \sU{\mathsf{U}}

\def \sZ{\mathsf{Z}}

\usepackage{xcolor}
\usepackage{bbm}

\usepackage{tikz}
\usetikzlibrary{shapes, positioning, calc}

\usepackage{parskip}
\usepackage{titlesec}

\titlespacing\section{0pt}{16pt plus 4pt minus 2pt}{6pt plus 2pt minus 2pt}
\titlespacing\subsection{0pt}{16pt plus 4pt minus 2pt}{6pt plus 2pt minus 2pt}
\titlespacing\subsubsection{0pt}{16pt plus 4pt minus 2pt}{6pt plus 2pt minus 2pt}

\titleformat{\section}[hang]{\normalfont\Large\bfseries}{\thesection}{0.5em}{}[]
\titleformat{\subsection}[hang]{\normalfont\Large\bfseries}{\thesubsection}{0.5em}{}[]
\titleformat{\subsubsection}[hang]{\normalfont\large\bfseries}{\thesubsubsection}{0.5em}{}[]

\ifpdf
\hypersetup{
  pdftitle={A Primer on Variational Inference for Physics-Informed Deep Generative Modelling},
  pdfauthor={A. Glyn-Davies, A. Vadeboncoeur, OD. Akyildiz, I. Kazlauskaite, M. Girolami }
}
\fi

\usepackage[acronym]{glossaries}
\newacronym{wrm}{WRM}{weighted residual method}
\newacronym{fe}{FE}{finite element}
\newacronym{uq}{UQ}{uncertainty quantification}
\newacronym{pinn}{PINN}{physics-informed neural networks}
\newacronym{ml}{ML}{machine learning}
\newacronym{phi-ml}{Phi-ML}{physics-informed machine learning}
\newacronym{vi}{VI}{variational inference}
\newacronym{bip}{BIP}{Bayesian inverse problem}
\newacronym{pde}{PDE}{partial differential equation}
\newacronym{ode}{ODE}{ordinary differential equation}
\newacronym{sde}{ODE}{stochastic differential equation}
\newacronym{elbo}{ELBO}{evidence lower bound}
\newacronym{vae}{VAE}{variaitonal auto-encoder}
\newacronym{kl}{$\mathrm{KL}$}{Kullback–Leibler}
\newacronym{mcmc}{MCMC}{Markov chain Monte Carlo}
\newacronym{em}{EM}{Expectation Maximisation}
\newacronym{dgp}{DGP}{deep generative prior}
\newacronym{gan}{GAN}{generative adversarial network}

\def \cZ {\mathcal{Z}}
\def \cU {\mathcal{U}}
\def \cY {\mathcal{Y}}
\def \cL {\mathcal{L}}

\def \cQ {\mathcal{Q}}
\def \cV {\mathcal{V}}
\def \cP {\mathcal{P}}
\def \cN {\mathcal{N}}

\def \vr {\mathbf{r}}
\def \vu {\mathbf{u}}
\def \vx {\mathbf{x}}
\def \vy {\mathbf{y}}
\def \vz {\mathbf{z}}

\def \vw {\mathbf{w}}

\def \md {\mathrm{d}}
\def \KL {\mathrm{KL}}

\def \bR {\mathbb{R}}
\def \bE {\mathbb{E}}

\def \argmin {\mathrm{arg\, min}}
\def \argmax {\mathrm{arg\, max}}

\usepackage{mathtools}

\title{{\scshape{A Primer on Variational Inference for Physics-Informed Deep Generative Modelling}}}

\newcommand{\cblue}{\textcolor{blue}}

\author[$\cblue{\dagger} $]{Alex Glyn-Davies$^\ast$}
\author[$\cblue{\dagger}$]{Arnaud Vadeboncoeur\footnote{Equal contributions, Corresponding author: Arnaud Vadeboncoeur (av537@cam.ac.uk).}}
\author[$\cblue{\natural}$]{O. Deniz Akyildiz}
\author[$\cblue{\ddagger}$]{\\ Ieva Kazlauskaite}
\author[$\cblue{\dagger} \S$]{Mark Girolami}

\affil[$\cblue{\dagger}$]{Department of Engineering, University of Cambridge}
\affil[$\cblue{\natural}$]{Department of Mathematics, Imperial College London}
\affil[$\cblue{\ddagger}$]{Department of Statistics, London School of Economics and Political Science}
\affil[$\cblue{\S}$]{The Alan Turing Institute}

\date{}
\date{\small \textbf{Keywords}: Deep Learning, Physics-Informed, Variational Inference, Generative Model, PDE}
\begin{document}
\setstretch{1.}

\maketitle

\begin{abstract}
\Gls*{vi} is a computationally efficient and scalable methodology for approximate Bayesian inference. It strikes a balance between accuracy of uncertainty quantification and practical tractability. It excels at generative modelling and inversion tasks due to its built-in Bayesian regularisation and flexibility, essential qualities for physics related problems. 
For such problems, the underlying physical model determines the dependence between variables of interest, which in turn will require a tailored derivation for the central VI learning objective. 
Furthermore, in many physical inference applications this structure has rich meaning and is essential for accurately capturing the dynamics of interest.
In this paper, we provide an accessible and thorough technical introduction to VI for  forward and inverse problems, guiding the reader through standard derivations of the VI framework and how it can best be realized through deep learning. We then review and unify recent literature exemplifying the flexibility allowed by \gls*{vi}. 
This paper is designed for a general scientific audience looking to solve physics-based problems with an emphasis on uncertainty quantification.
\end{abstract}

\section{Introduction}
\noindent This paper serves as tutorial and review on  methodologies for inference related to physical problems  using \gls*{vi}.  We    introduce
 basic concepts and the   mathematical  formulations pertaining to the most relevant and important tools in the field. 
 {We first consider the modelling of physical systems with \glspl*{pde}.}
 We then present an overview of inverse problems through optimisation and Bayesian perspectives, and provide a detailed derivation of \gls*{vi}. Equipped with this knowledge we then review salient methods  in the literature
for solving physical inference problems with forward model and
{\gls*{wrm} -based \gls*{vi}.}

\textbf{Forward Problems} in physical modelling refer to the computation, simulation or estimation of the solution to a mathematical physics problem. These can come in a variety of forms such as agent-based models~\cite{janssen2006empirically}, data-driven models~\cite{kirchdoerfer2016data}, differential equations~\cite{finlayson2013method} and any number of combinations thereof.
In this work we focus on models which describe mechanistic understanding through differential equations. Broadly speaking these models describe the change in certain quantities of interest, such as heat, velocity, electric potential, with respect to time or space.
{ As such, these models are intrinsically linked to the setting in which they are considered, that is to say, initial conditions, boundary conditions, geometry, and other physical quantities.}
If multiple forward problems must be solved for different sets of parameters, classical numerical solvers can be computationally intractable. 
These multi-query problems often arise in contexts of \gls*{uq}
through methods such as Monte Carlo sampling, Taylor expansion and perturbation methods. Surrogate models may alleviate this computational burden~\cite{sullivan2015introduction}. 
{A classical example of surrogate models for forward problems are Gaussian Processes (GPs), which have inherent uncertainty quantification capabilities~\cite{kennedy2001bayesian}.}
Many learning models have been recently developed for surrogate modelling of PDEs with functional inputs such as deep operator networks (DeepONet) and Fourier Neural Operators (FNO)~\cite{li2021fourier, lu2019deeponet} however these models do not have built-in \gls*{uq} capabilities like~\cite{psaros2023uncertainty, lin2023b}.

\textbf{Inverse Problems} on the other hand,  aim to recover model parameters that gave rise to a set of observations,  i.e. inverting the forward problem.
Classic application fields include Computed Tomography~\cite{ramm2020radon};  cosmology~\cite{trotta2008bayes},
and geophysics~\cite{zhdanov2002geophysical}. 
When observations are noisy or sparse, the inverse problem is typically \textit{ill-posed}, meaning that many different model parameter values could have provided the same observations. Then, inverse problems require a form of \textit{regularisation} on the model parameters to provide unique solutions~\cite{benning2018modern}. 
{Point-estimate-based inversion generally do not seek UQ}~\cite{arridge2019solving}, while Bayesian methods recover distributions over parameters~\cite{stuart2010inverse}. 

\textbf{Variational Inference}
is a statistical framework which strikes a practical balance between computational costs and accuracy of \gls*{uq}~\cite{bishop2006pattern, mcgrory2007variational}. 
{
It relies on the optimisation of a statistical objective to provide uncertainty estimates in inference tasks~\cite{blei2017variational}.
}
There is a large variety of \gls*{vi} schemes 
with different advantages and limitations~\cite{knoblauch2019generalized}. {One of the most discernible advantages} of constructing \gls*{vi}-based inference schemes is to allow one to circumvent expensive \gls*{mcmc} sampling of intractable probability distributions which often arise in the statistical treatment of uncertainty relating to nonlinear models. As these nonlinear models are essential for capturing the physical structure of many scientific problems, \gls*{vi} methods have great potential in making \gls*{uq} for sciences computationally feasible. Furthermore, \gls*{vi} allows practitioners to construct computationally efficient frameworks with built-in conditional dependence structures reflecting the nature of the inferential task at hand~\cite{jordan1999introduction, wainwright2008graphical, pearl2014probabilistic}. This conditional dependence structure will often be represented as a Bayesian graphical model~\cite{barber2012bayesian, bishop2006pattern}. The ability to strictly enforce intricate dependencies between quantities of interest  -- such as in physics problems --  is precisely what gives rise to the {wide} variety of methods explored in this paper.

We structure the rest of the paper as follows: Section~\ref{sec:physics-and-inference} introduces the relevant mathematical background; forward problems are described in Section \ref{subsec:forward}; optimisation and Bayesian inference for inverse problems are covered in Section \ref{subsec:inverse}; \gls*{vi} methods are presented in Section \ref{subsec:VI}. {Section \ref{sec:pde-informed} reviews applications of these methods to physics-based generative modelling tasks found in the literature.} Applications are split into forward-model-based approaches  in Section \ref{subsec:forward_based_learning}, and residual-based learning in Section \ref{subsec:residual_based_learning}.

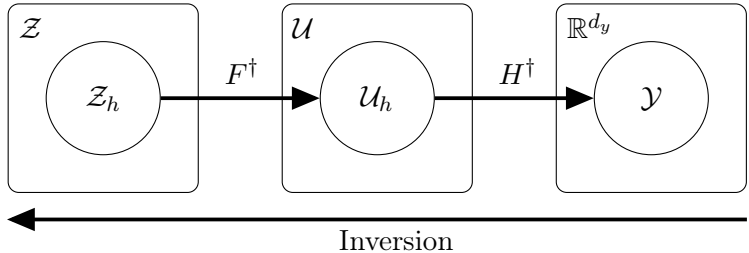
\begin{figure}
    \centering
    \begin{tikzpicture}[->]
         Define the rounded rectangle node style
        \tikzstyle{rounded rectangle}=[>=stealth, auto, rectangle, rounded corners, minimum height=2cm, minimum width=12cm]
        
        \tikzstyle{subspace}=[>=stealth, auto, draw, rectangle, rounded corners, minimum height=2.5cm, minimum width=2.5cm]
    
        \node[rounded rectangle] (rect) {};
    
        \node[circle, draw, minimum size=1.5cm, text centered] (node2) at ($(rect.west)!0.5!(rect.east)$) {$\cU_h$};
        \node[subspace] (ssu)  at ($(rect.west)!0.5!(rect.east)$) {};
        
        \node[circle, draw, minimum size=1.5cm, text centered] (node1) at ($(rect.west)!0.2!(rect.east)$) {$\cZ_h$};
        \node[subspace] (ssy)  at ($(rect.west)!0.2!(rect.east)$) {};
                
        \node[circle, draw, minimum size=1.5cm, text centered] (node3) at ($(rect.west)!0.8!(rect.east)$) {$\cY$};
        \node[subspace] (ssz)  at ($(rect.west)!0.8!(rect.east)$) {};
    

        \node[above right=-0.6cm and -2.5cm of ssu.north east, anchor=south west] {$\cU$};
        \node[above right=-0.6cm and -2.5cm of ssy.north east, anchor=south west] {$\cZ$};        
        \node[above right=-0.6cm and -2.5cm of ssz.north east, anchor=south west] {$\bR^{d_y}$};

        \node[above right=-3cm and -10cm of ssz.north east, anchor=south west] (lineend) {};
        \node[above right=-3cm and -0cm of ssz.north east, anchor=south west] (linestart) {};

        \path [every node/.style={font=\sffamily\small}]
        (node1) edge[thick, line width=0.5mm, bend left=0] node [left] {} (node2)
        (node2) edge[thick, line width=0.5mm, bend left=0] node [left] {} (node3)
        (linestart) edge[thick, line width=0.5mm, bend left=0] node [left] {} (lineend);

        \node[above right=-1.2cm and -7.cm of ssz.north east, anchor=south west] {$F^\dagger$};
        \node[above right=-1.2cm and -3.4cm of ssz.north east, anchor=south west] {$H^\dagger$};
        \node[above right=-3.4cm and -5.5cm of ssz.north east, anchor=south west] {Inversion};


    \end{tikzpicture}
    \caption{A depiction of the three spaces of inferential interest, the observation space $\cY$, the discretised solution space $\cU_h$, and the discretised parameter space $\cZ_h$.
    More specifically, we have an observation $\vy\in \cY\subseteq\bR^{d_y}$, a solution $u_h\in\cU_h\subset\cU$, and a parameter $z_h\in\cZ_h\subset\cZ$.
    }
    \label{fig:spaces}
\end{figure}

\section{Physics and Inference}
\label{sec:physics-and-inference}
In this section we introduce and elaborate on the core concepts and tools required to build variational inference schemes for the physical sciences. 
In Fig.~\ref{fig:spaces} we show a depiction of the mathematical spaces that describe the three main quantities of inferential interest: parameter, solution, observation, which we denote as $z\in \cZ, u\in \cU, y\in \cY$ respectively.
In the following sections we denote the {finite-dimensional} representations of the parameter and solution as $z_h\in \cZ_h, u_h\in \cU_h$ respectively, where the subscript $h$ is a parameter describing the degree of discretisation. This means we only consider spaces of solution and parameters which are finite-dimensional, 
hence they have already been discretised. Rigorous mathematical treatment of inference schemes over functions, which are infinite-dimensional, is of great value but beyond the scope of this paper~\cite{stuart2010inverse, gine2021mathematical}.

\subsection{Forward Problems}
\label{subsec:forward}
We describe a generic forward model through a numerical scheme which relates the discretised physical setup, $z_h\in\cZ_h$ to the realization of the physical process across time and space, which we call the solution and is denoted as $u_h\in\cU_h$. The forward model is a mapping from a particular setup to the solution associated to that setup described as $F^\dagger:\cZ_h\rightarrow\cU_h$. The use of ``$\dagger$'' refers to the near exact numerical realization of the differential equations of interest, and we will see later how this might be approximated by a parametrised -- less expensive to evaluate -- surrogate model.

To discuss \glspl*{pde} in more detail we choose a canonical example, the Poisson problem. It describes a variety of steady-state diffusive physical systems such as heat, electric potential, ground water flow etc. A function is said to be a solution to this problem if it satisfies, for some physical domain $\Omega$,
\begin{subequations}\label{eq:poisson}
\begin{alignat}{2}
    \nabla \cdot( z(x) \nabla u(x)) &= f(x),\quad&&\mathrm{for}\, x\in \Omega,\\
    u(x) &= 0,&&\mathrm{for}\, x\in \partial\Omega,
\end{alignat}
\end{subequations}
where $\partial\Omega$ denotes the boundary of $\Omega$. 
The problem stated in this form is not  amenable to numerical computation as $u$ is currently an infinite-dimensional object and it must be discretised.
How we represent this function $u\in\cU$ and how it relates to~\eqref{eq:poisson} is given by the particular numerical scheme in use.

We look at the discretisation of solution fields and PDE operators through the lens of the \gls*{wrm}~\cite{finlayson2013method} which encompasses most spatial discretisation schemes such as \gls*{fe}, spectral methods, finite difference and \glspl*{pinn}. The advantage of taking this perspective on numerical discretisation for \gls*{ml} is that inference schemes can be constructed independently of the $\textit{particular}$ \gls*{wrm} method in use, hence these can be swapped out with ease. 
To write out the weighted residual methods we first specify the residual function 
\begin{align}
    R(u, z, f, x) = \nabla \cdot( z(x) \nabla u(x))  - f(x).
\end{align}
Choosing a set of weight functions $\{v_i\}_{i=1}^{d_r}$ with $v_i\in\cV$ we can test the residual 
\begin{align}\label{eq:res}
    r_i = \int_\Omega v_i(x) R(u, z, f, x)\, \md x = \int_\Omega  v_i(x) \left(\nabla \cdot( z(x) \nabla u(x))  - f(x)\right)\, \md x.
\end{align}
Collecting $\vr=\{r_i\}_{i=1}^{d_r}$ discretises the action of the differential operator on the solution $u$.
One can then  use integration by parts on~\eqref{eq:res} if the test functions are differentiable {to} obtain the \textit{weak form} of the Poisson equation, 
\begin{align}
    r_i = \int_{\partial\Omega}  v_i(x) ( z(x) \nabla u(x)) \cdot \hat{n}(x)\, \md x - \int_\Omega \nabla v_i(x) \cdot ( z(x) \nabla u(x))\, \md x  - \int_\Omega f(x)\, \md x.
\end{align}
Various other Galerkin-type methods can be designed by varying the choice of test and trial functions. By choosing $v_i=\phi_i$ (implying $v\in\cV^h=\cU^h$ and $\cV_h=\mathrm{span}\{\phi_i\}_{i=1}^{N_u}$ and for this problem choosing $\phi_i$ to be hat functions) we obtain a Bubnov-Galerkin method~\cite{reddy1993introduction}.
Working with such weak forms has notable advantages, mainly it reduces the differentiability requirements on the trial function as a derivative order is passed over to the test function. 
Linear approximants can be represented with the following basis function expansion $u_h(x) = \sum_{i=1}^{N_u}[\vu]_i\,\phi_i(x)$ where $u_h\in\cU_h$, $\vu\in \sU$ are the coefficients, and $\phi_i$ are the basis functions.
When constructing inference schemes we can now use $\vu$ in lieu of $u_h$. Similarly, we can replace $z\in\cZ$ -- which in this particular example is a function -- with a finite-dimensional discretisation $z_h\in\cZ_h$ which in turn can be expressed with an expansion as $z_h(x) = \sum_{i=1}^{N_z}[\vz]_i\,\psi_i(x)$ and summarized as $\vz\in\sZ$.
We denote the chosen mapping from coefficients $\vz, \vu$ to interpolants $z_h, u_h$ as $\pi_z(\vz)=z_h$, $\pi_u(\vu)=u_h$ respectively.
Residuals like these can be efficiently computed in a GPU-efficient manner using array-shifting~\cite{akyildiz2024efficient} or convolutions~\cite{zhu2019physics}. We note that a variety of variational formulations such as the Ritz method or energy functionals are amenable to equivalent residual formulations as in~\eqref{eq:res}~\cite{leissa2005historical}.

{
\glspl*{pinn} are neural network based methods for approximating the solution to differential equations~\cite{raissi2019physics}.
Many of these methods can be obtained by taking $u_h$ to be a nonlinear approximant as a neural network. A typical form }
is $u_h(x) = T_L \circ\hdots\circ T_0(x)$ where $T_i(x) = \boldsymbol{\sigma}_i(W_i\, x + b_i)$ where $\boldsymbol{\sigma}_i, W_i,\,b_i$ are the layers' activation function, weight matrix  and bias vector, respectively, and choosing $v_i(x) = \delta(x_i - x)$ where $\delta$ is the Dirac-delta function and $x_i$ are collocation points. For these \gls*{pde} solvers, the solution representation for inference is $\vu=\{u_h(x_i)\}_{i=1}^{N_u}$. It is to be noted that when using this kind of approach, we no longer make use of the weak form. Neural network approximants may still be used with the variational form~\cite{kharazmi2021hp}. 
{
For further reading on this topic we refer readers to ~\cite{lu2021deepxde, karniadakis2021physics, cai2021physics}.
}

The treatment of boundary conditions depends on the specific \gls*{wrm} method in use; \gls*{fe}-based methods typically use {boundary-respecting} meshes and the weak form naturally includes other boundary conditions; \glspl*{pinn}-style methods can either include an additive boundary loss term to the residual or enforce certain types of boundary conditions through certain manipulations of $u_h$~\cite{sukumar2022exact}.
To numerically solve the PDE means to find $u_h$ such that the residual vector {$\vr\approx0$}, {within a pre-defined tolerance}. 
In the case of the \gls*{fe} method for linear \glspl*{pde}, a system of sparse linear equations can be setup which can be directly solved using linear solvers, but the residual formulation may still be implemented as is often done in the case of~\glspl*{pinn}.

\subsection{Inverse Problems}
\label{subsec:inverse}
{Inversion methods map elements of $\cY$ to points or distributions in $\cU$ or $\cZ$~\footnote{{In this work, we consider data assimilation, i.e. recovering trajectories/solutions from observational data, to be a subset of inversion tasks.}}.}
That is, we either wish to recover the full solution from observations, or the parameters from observations. 
We find it appropriate to separate the full mapping between parameter-to-observation, denoted $G^\dagger$, into the mapping from parameter-to-solution, $F^\dagger$ (forward model), and the mapping from solution-to-observation, $H^\dagger:\cU_h\rightarrow \cY$ (observation model). Here, the ``$\dagger$'' denotes the ``true mapping'' to distinguish from settings where we might try and learn this map. The full parameter-to-observation map can be written as $G^\dagger(z_h) = (H^\dagger\circ F^\dagger) (z_h)$, the composition of the forward and observation maps.
\subsubsection{{Point Estimate Inversion}}
{If one is not interested in recovering uncertainty over model parameters given some data, point estimate inversion may be used.
}
Inversion schemes rely on the combination of a data-fit term and a regularisation term.
As most inverse problems of interest are ill-posed, the quality of the estimated quantities from applying inversion schemes is tied to the quality of the regularisation imposed.
A classic approach to the regularisation of inverse problems is the Tikhonov approach~\cite{stuart2010inverse, arridge2019solving, benning2018modern}
\begin{align}\label{eq:tikh}
    \vz^\star = \underset{\vz\in\sZ}{\argmin}\quad \frac{1}{2}\|\vy - (H^\dagger\circ F^\dagger\circ\pi_z)(\vz)\|^2 + \frac{\beta}{2}\|\pi_z(\vz)\|^2,
\end{align}
where $F^\dagger$ is the forward model and $\beta$ controls the strength of the bias towards $z_h$ estimates that are small in the chosen norm. 
{We note other forms of regularization are possible, such as total variation~\cite{chan2005recent}, sparsity promoting $\ell_1$ regularization~\cite{tibshirani1996regression} and regularizing operators~\cite{zhdanov2002geophysical}.
}
Alternative perspectives on inverse problems for physical system use the regularisation term to impose physical knowledge. These methods estimate the parameter of interest as 
\begin{align}
    \vz^\star = \underset{\vz\in\sZ}{\mathrm{arg min}} \;\underset{\vu\in \sU}{\mathrm{min}}\quad \|\vy - (H^\dagger\circ \pi_u)(\vu)\|^2 + \beta\|\vr(\pi_u(\vu); \pi_z(\vz))\|^2{,}
\end{align}
where $\beta$ now controls the tradeoff between the data-fit and the physics regularisation.  In practice, the parameter $\beta$ is {often} manually tuned. 
Taking $u_h$ as the output of a \glspl*{pinn} and the \gls*{wrm} used for computing $\vr \in \mathbb{R}^{d_r}$ to be a collocation method where the test functions are Diracs 
recovers a \glspl*{pinn}-style parameter inversion method. We note one can choose $u_h$ to be a \gls*{fe} expansion with a weak form result computation. 
An interesting development on these methods is to formulate the combined objectives in terms of a bilevel optimisation problem~\cite{holler2018bilevel} which eliminates the need to balance the physics residual with the data-fit term.

\subsubsection{Bayesian Inverse Problems}
{Recovering a point estimate of the solution may be insufficient for many applications.}
\Glspl*{bip} provide an alternative approach through the probabilistic framework of Bayes' theorem that offers a unifying framework, \gls*{uq} and some theoretical insights into the posterior consistency of the recovered solution. 
Bayes' theorem, given as
\begin{align}\label{eq:bayes_post}
    p(\vz|\vy) = \frac{p(\vy|\vz)p(\vz)}{p(\vy)},\quad \text{where} \quad  
    p(\vy) = \int p(\vy|\vz)p(\vz)\, \md\vz,
\end{align} 
allows one to derive the full posterior distribution over the model parameters $\vz$ given the observed data $\vy$. This approach combines the likelihood $p(\vy | \vz)$, derived from the data-generating model, and the prior distribution $p(\vz)$ as the regulariser, offering a direct parallel to the {point-estimate-based} approach. 
The model evidence, $p(\vy)$, also known as the marginal likelihood, which appears in \eqref{eq:bayes_post}, is often intractable. Hence the need for methods that do not require normalized probability densities such as \gls*{mcmc} or Bayes \gls*{vi}.
Note that the point estimate recovered using the optimisation approach is typically the Maximum A Posteriori (MAP) estimate (as in Eq.~\ref{eq:tikh}) where additive zero-mean Gaussian noise on the observations leads to a Gaussian likelihood.
For typical physical systems, the mapping from parameter to observation can be expressed as $G =  (H^\dagger\circ F^\dagger\circ\pi_z)$.
We consider a set of observations that arise as independent and identically distributed (i.i.d.) 
\begin{align}\label{eq:obs_model}
    \vy = G(\vz) +  \boldsymbol{\epsilon}, \quad \boldsymbol{\epsilon} \sim \cN(0, \Gamma),
\end{align}
where $\Gamma$ is the symmetric positive-definite noise covariance. The observation model~\eqref{eq:obs_model} results in a Gaussian likelihood $p(\vy|\vz)=\cN(\vy; G(\vz), \Gamma)$.
\subsection{Variational Inference}
\label{subsec:VI}
At its core, \gls*{vi} poses statistical inference as an optimisation problem by minimizing a data-informed \textit{regularised loss} over a \textit{variational family} of distributions. Abstractly, we seek
\begin{align}
    q^\star(\vz) \in \underset{q\in\cQ(\sZ)}{\argmin}\; J(q(\vz);\,\vy),
\end{align}
where $\cQ(\sZ)\subseteq\cP(\sZ)$ 
is the variational family -- a subset of all possible probability measures on $\sZ$. To realize this approach we typically choose $\cQ(\sZ)$ to have a parametric form with parameters $\phi$. The variational approximation $q_\phi(\vz)$ (with $\phi$ being the mean and covariance for Gaussian approximations, for example) is then parametrised by $\phi$ and loss is minimised with respect to $\phi$. In some cases, closed forms of the updates on $\phi$ can be derived, but in many modern application one resorts to gradient descent schemes. The choice of loss function $J(\,\cdot\,;\vy)$ is crucial and determines the object recovered by the method. We next discuss two pertinent concepts: Bayes~\gls*{vi} and probabilistic generative models.

\subsubsection{Bayes Variational Inference}
Bayes \gls*{vi} is the optimisation formulation of Bayes theorem. It performs inference with a principled balance between data-fit and prior knowledge and recovers a probability distribution over model parameters.
The loss function for Bayes \gls*{vi} is based on the \gls*{kl} divergence
\begin{align}
    D_\KL(q(\vz)||p(\vz)) = \bE_{q(\vz)}\left[\log\frac{q(\vz)}{p(\vz)}\right],
\end{align}
given absolute continuity between $q$ and $p$, meaning $q$ assigns zero probability to sets for which $p$ also assigns zero probability. 
The KL divergence quantifies the difference between two probability distributions. 
Bayes \gls*{vi} aims to minimise the \gls*{kl} divergence between the true posterior $p(\vz|\vy)$, and the variational approximation $q_{\phi}(\vz)$, parametrised by $\phi$. To derive the objective function, {we write out the \gls*{kl} divergence, before applying Bayes' theorem and simplifying}
 \begin{align}
     D_\KL(q_{\phi}(\vz)||p(\vz|\vy))&{=
     \bE_{q_\phi(\vz)}\left[\log\frac{q_\phi(\vz)}{p(\vz|\vy)}\right] = \bE_{q_\phi(\vz)}\left[\log\frac{p(\vy)q_\phi(\vz)}{p(\vy|\vz)p(\vz)}\right],}\nonumber\\
     &=\log p(\vy) - \bE_{q_{\phi}(\vz)}\left[ \log p(\vy|\vz)  \right] + \bE_{q_{\phi}(\vz)}\left[ \log \frac{q_{\phi}(\vz)}{p(\vz)}  \right].\label{eq:poseterior_KL}
\end{align}
As $p(\vy)$ does not depend on the variational approximation $q_{\phi}(\vz)$~\cite{sanz2023inverse}, 
minimizing $D_\KL(q_\phi(\vz)||p(\vz|\vy))$ is equivalent to minimizing
\begin{align}\label{eq:J_Bayes_VI}
    J(\phi;\vy) \coloneqq \bE_{q_{\phi}(\vz)}\left[ - \log p(\vy|\vz)  \right] +D_\KL(q_{\phi}(\vz)||p(\vz)).
\end{align} 
In this form $J(\phi;\vy)$ avoids the expensive computation of the model evidence $p(\vy)$ and is directly minimizing the KL divergence between the variational approximation and the Bayes' posterior.
Seeking $\phi^\star = \argmin_\phi\;J(\phi;\vy)$, yields a Bayes \gls*{vi} approximation to the posterior.
In practice, the expectations in \eqref{eq:J_Bayes_VI} are approximated via Monte Carlo using samples $\vz^{(s)} \sim q(\vz)$, $s = 1, \dots, S$~\cite{ranganath2014black}.  

\subsubsection{Probabilistic Generative Models}

Probabilistic generative models are defined by a joint distribution $p_{\theta}(\vz, \vy)$, parametrised by $\theta$ which are to be estimated from the observed data. In order to learn the generative model, these parameters are typically estimated via maximisation of the Bayesian model evidence, $p_{\theta}(\vy) = \int p_{\theta}(\vz, \vy)\md \vz$, which now depends on $\theta$.   
Methods in variational inference, such as \glspl*{vae}~\cite{kingma2013auto} will often combine estimation of generative model parameters, with the variational approximation of the posterior $q_{\phi}(\vz)${, where, in general, the exact posterior $p_\theta(\vz|\vy) = p_\theta(\vz,\vy)/\int p_\theta(\vz,\vy) \md\vz$ cannot be evaluated due to the intractable normalisation constant arising from the complex generative model structure}. In such cases, the joint estimation of parameters $\left\{\phi, \theta\right\}$ is required. Taking the prior $p(\vz)$ as fixed, and the likelihood $p_{\theta}(\vy|\vz)$ as the parametrised model, we can rearrange \eqref{eq:poseterior_KL}, to obtain an expression for the log-marginal likelihood, 
\begin{align}\label{eq:log_ml_theta}
     \log p_{\theta}(\vy) = D_\KL(q_{\phi}(\vz)||p_{\theta}(\vz|\vy)) + \bE_{q_{\phi}(\vz)}\left[ \log p_{\theta}(\vy|\vz)  \right] - D_\KL(q_{\phi}(\vz)||p(\vz)),
\end{align}
which is intractable due to the evaluation of the posterior {in the first right-hand term}, but can be bounded from below due to the non-negativity of the \gls*{kl}
\begin{align}
     \log p_{\theta}(\vy) \geq \bE_{q_{\phi}(\vz)}\left[ \log p_{\theta}(\vy|\vz)  \right] - D_\KL(q_{\phi}(\vz)||p(\vz)) \coloneqq \mathcal{L}(\phi,\theta;\vy).
\end{align}
Here, $\mathcal{L}$ is known as the \gls*{elbo}, and in practice is maximised via gradient-based stochastic optimisation schemes, using Monte Carlo to estimate expectations.
For optimisation the objective is defined in terms of both $\phi, \theta$ as the negative \gls*{elbo}, $J(\phi, \theta;\vy)\coloneqq -\mathcal{L}(\phi,\theta;\vy)$, where optimal parameters minimise this objective $\phi^\star, \theta^\star = \argmin_{\phi, \theta}\;J(\phi, \theta;\vy)$.

{
We note that the \gls*{elbo} is often derived via Jensen's inequality (see e.g.~\cite{zhang2018advances}), which applies to concave transformations of expectations, and for the natural log reads $\log(\bE[X])\geq \bE[\log(X)]$~\cite{jordan1999introduction}, and is applied for \eqref{eq:jensen-applied} below 
    \begin{align}
        \log p_\theta(\vy) &= \log \left( \int p_\theta(\vz, \vy)  \md \vz \right) =  \log \left( \int \frac{p_\theta(\vz, \vy)}{q_{\phi}(\vz)} q_{\phi}(\vz) \md \vz \right)\\
                           &= \log \left( \bE_{q_{\phi}(\vz)}\left[\frac{p_\theta(\vz, \vy)}{q_{\phi}(\vz)}\right] \right) \geq  \bE_{q_{\phi}(\vz)}\left[\log\frac{p_\theta(\vz, \vy)}{q_{\phi}(\vz)}\right]\label{eq:jensen-applied}\\
                           &= \bE_{q_{\phi}(\vz)}\left[\log p_\theta(\vy|\vz)\right] - D_\KL (q_{\phi}(\vz)||p(\vz)) = \cL(\vy; \phi,\theta).
    \end{align}
It is important to note that since the \gls*{kl} term dropped from \eqref{eq:log_ml_theta} depends on $\theta$, $\mathcal{L}$ is a \textit{lower bound}, whereas in \eqref{eq:poseterior_KL} the objective is directly minimising the posterior \gls*{kl} without approximation (as $\log p(\vy)$ does not depend on  $\theta$).
}

The \gls*{elbo} is used for unsupervised learning in \glspl*{vae}, which are probabilistic generative models defined by an encoder and decoder. The encoder is a conditional distribution $q_{\phi}(\vz|\vy)$ which, intuitively, \textit{encodes} a data point $\vy$ into the latent space $\mathsf{Z}$ by returning a probability distribution over it (rather than a fixed embedding). Similarly, the probabilistic decoder $p_{\theta}(\vy|\vz)$ is a probability measure for fixed $\vz$ and $\theta$, meaning that the decoder returns a distribution over the data $\vy$ given the latent vector $\vz$.

The latent space is typically low-dimensional, forcing the model to learn parsimonious representations of the data, and is regularised by a (often simple) prior distribution, e.g. $p(\vz) = \cN(0,\mathrm{I})$. 
Both $q_{\phi}$ and $p_{\theta}$, in general, are parametrised with neural networks. For a dataset $\mathcal{D} = \{\vy^{(n)}\}_{n=1}^{N}$, and assuming i.i.d. observations such that the log likelihood decomposes as $\log p_{\theta}(\vy^{(1:N)}) = \sum_{n=1}^N \log p_{\theta}(\vy^{(n)})$, we can write the log marginal likelihood as
\begin{align}
    \label{eq:vae_elbo}
    \log p_{\theta}(\vy^{(1:N)}) &\geq
    \sum_{n=1}^N \underbrace{\mathbb{E}_{q_\phi(\vz|\vy^{(n)})}\left[\log p_\theta(\vy^{(n)}|\vz)\right]}_{\text{reconstruction error}} - \underbrace{D_{\KL}(q_\phi(\vz|\vy^{(n)}) || p(\vz))}_{\text{regularisation}}
    \eqqcolon \sum_{n=1}^N\mathcal{L}(\vy^{(n)}; \theta, \phi).
\end{align}
For large datasets, one often uses a \textit{mini-batch}, $B \subseteq \mathcal{D}$, of the dataset per gradient step, giving an approximate minimisation objective $J(\theta, \phi; \vy^{(1:N)}) \coloneqq -\frac{N}{|B|}\sum_{n\in B}\mathcal{L}(\vy^{(n)}; \theta, \phi)$. As we approximate this lower bound stochastically through Monte Carlo, our objective is a 'doubly-stochastic' approximation to the true \gls*{elbo} which is found to improve learning~\cite{kingma2013auto}.  
If we now choose $q_\phi(\vz|\vy^{(n)})=\cN(\vz; m_\phi(\vy^{(n)}), C_\phi(\vy^{(n)}))$ and $p_\theta(\vy^{(n)}|\vz)=\cN(\vy^{(n)};G_\theta(\vz), C_\eta)$ with $m_\phi(\cdot)$, $C_\phi(\cdot)$, $G_\theta(\cdot)$, being neural networks, we obtain the classic \gls*{vae}. The choice of prior distribution affects the latent regularisation, and is typically chosen as a standard Gaussian, $p(\vz) \sim \cN(\vz; 0, \mathrm{I})$.

{
A practical consideration when training \glspl*{vae} is the computation of the loss function's gradient with respect to the VI parameters $\nabla_\phi J(\theta, \phi;\vy^{(1:N)})$, which requires gradient backpropagation through the Monte Carlo sampled latent variables $\vz^{(i)} \sim q_{\phi}(\vz | \vy^{(n)})$. In order to facilitate the gradient backpropagation, practitioners employ the so-called ``reparameterisation-trick''~\cite{kingma2013auto}, which defines the latent random variable as a \textit{differentiable} transformation of the variational parameters, and a noise random variable, $\epsilon \sim p(\epsilon)$. For the Gaussian variational posterior above, this can be done by first sampling $\boldsymbol{\epsilon}\sim \cN(\boldsymbol{\epsilon}; 0, \mathrm{I})$, then transforming these to samples from the variational posterior as $\vz^{(i)} = m_{\phi}(\vy^{(n)}) + L_\phi(\vy^{(n)})\odot \boldsymbol{\epsilon}$, where $L_{\phi}(\vy^{(n)})$ is the Cholesky factor of $C_\phi(\vy^{(n)}) = L_\phi(\vy^{(n)})L_\phi(\vy^{(n)})^\top$. 
}

Constructing more expressive variational approximations can be achieved through normalizing flows~\cite{dinh2022density, rezende2015variational}. A complicated distribution is modelled as a series of invertible transformations of a simple reference distribution, e.g. $p(\vw) = \mathcal{N}(0,\mathrm{I})$. More explicitly, ${\vw^{(i)}}\sim p(\vw),\, {\vz^{(i)}}\sim q_\phi(\vz)$, where ${\vz^{(i)}} = f_{\phi} ({\vw^{(i)}})$ .
The density for $q_\phi(\vz)$ is computed through the change of variable formula $q_\phi(\vz) = p(f_\phi^{-1}(\vz)) \det \vert \partial_\vz f^{-1}_\phi(\vz)\vert$. Conditional normalising flows extended the normalising flow method to learn conditional densities, i.e. $q_{\phi}(\vz|\vy)$ similar to the encoder for a \gls*{vae}. Normalising flows have the benefit over \glspl*{vae} of being \textit{invertible} transformations, but as a result are constrained to having the same latent dimension as that of the data, so do not benefit from dimensionality reduction.

\section{Physics-Informed Generative Models}
\label{sec:pde-informed}
We now delve into salient works taken from the literature which best exemplify the flexibility and versatility of \gls*{vi} for physics.
In what follows we cast the central \gls*{vi} objective of selected works in a notation consistent with the previously presented material. This should be interpreted as a paraphrasing of the methods in the referenced works to help the reader best understand their differences and similarities. Particular implementation details such as precise residual computations or variational forms will vary.

\subsection{Forward-Model-based Learning}\label{subsec:forward_based_learning}
In this section we describe inverse problem methodologies that embed the forward model into the probabilistic generative model. It is assumed the forward model (while still potentially expensive) can be evaluated for a given input $\vz$ -- outputting a corresponding $\vy$ -- and the dataset is a collection of these physical model input-output pairs, $\mathcal{D} = \{\vz^{(n)}, \vy^{(n)}\}_{n=1}^{N}$. For a probabilistic generative model, this amounts to sampling from the joint distribution $p(\vz,\vy)\propto p(\vz)p(\vy|\vz)$.
In this setting, the likelihood describes a probabilistic forward map, as determined by the true forward model $G^\dagger(\cdot)$ and an assumed noise model, e.g. \eqref{eq:obs_model}. The central goal of these methodologies is to learn a variational approximation $q_{\phi}(\vz|\vy)$, that once trained, provides a calibrated posterior estimate over parameters for a previously unseen data-point. 

\subsubsection{Supervised VAEs for Calibrated Posteriors}
This class of models are for \textit{supervised} learning problems -- meaning we have access to input-output pairs. This allows for the use of the \textit{forward} \gls*{kl}, $D_{\KL}(p(\vz|\vy)||q_{\phi}(\vz|\vy))$ in the objective, as opposed to the mode-seeking reverse \gls*{kl}. The estimation of the mean-seeking forward \gls*{kl} requires an expectation with respect to the true posterior, which is unavailable to us. However, the average over the data distribution can be computed using samples from the joint distribution $p(\vz,\vy)$ via 
\begin{align}\label{eq:forward_KL_avg}
    \mathbb{E}_{p(\vy)} \left[ D_{\KL}(p(\vz|\vy)||q_{\phi}(\vz|\vy))\right] = \mathbb{E}_{p(\vz, \vy)}\left[-\log q_\phi(\vz|\vy)\right].
\end{align}
This approach is used in \cite{siahkoohi2023reliable} to learn an amortized variational approximation with sampled input-output pairs, computed via the true forward model by pushing prior samples $\vz^{(n)}\sim p(\vz)$ through the forward model, and sampling $\vy^{(n)} \sim \mathcal{N}(G^{\dagger}(\vz^{(n)}), \sigma^2 \mathrm{I})$. A conditional normalising flow provides the variational approximation $q_{\phi}(\vz|\vy) = \mathcal{N}(f_{\phi}^{-1}(\vz;\vy); 0,\mathrm{I}) \det \vert\partial_\vz f_{\phi}^{-1}(\vz;\vy) \vert$, mapping data to the latent space, and acting as a surrogate. The forward \gls*{kl} averaged over the data distribution, and \eqref{eq:forward_KL_avg} is the objective to learn the conditional normalising flow as 
\begin{align}
    \phi^{\star} = \underset{\phi}{\argmin} \; J(\phi;\vy), \quad
    J(\phi;\vy) = \mathbb{E}_{p(\vz, \vy)}\left[\frac{1}{2}\Vert f_{\phi}^{-1}(\vz;\vy)\Vert^2_2 - \log \det\vert\partial_\vz f_{\phi}^{-1}(\vz;\vy) \vert \right].
\end{align}
The posterior given an unseen data-point is then computed by sampling $\vw\sim\mathcal{N}(0,\mathrm{I})$ and pushing through the trained conditional normalising flow $f_{\phi^\star}(\vw;\vy)$ which approximately samples from $p(\vz|\vy)$.

In~\cite{gohSolvingBayesianInverse2022}, the decoder of a \gls*{vae} is replaced by the known physical forward model which acts to physically regularise the problem. Data is assumed to be observed under some known noise model $\vy \sim \mathcal{N}(G^\dagger (\vz) + m_{\epsilon}, C_{\epsilon})$
, which can include a bias through the mean $m_{\epsilon}$. Input-output pairs are used to learn an amortized variational posterior with mean $m_{\phi}(\cdot)$, and covariance square-root $C_{\phi}^{1/2}(\cdot)$ parametrised by neural networks, yielding $q_{\phi}(\vz|\vy) = \mathcal{N}(m_{\phi}(\vy), C_{\phi}(\vy))$. The Jensen-Shannon divergence 
, which is parametrised by $\alpha \in [0,1]$, interpolates between the forward ($\alpha=0$) and reverse ($\alpha=1$) \gls*{kl}. The form of this divergence between $q\coloneqq q(\vz)$ and $p\coloneqq p(\vz)$ is
\begin{align}
    \mathrm{JS}_{\alpha}(q||p) = 
    \alpha D_{\KL}(q|| (1-\alpha)q + \alpha p) 
    + (1-\alpha)D_{\KL}(p||(1-\alpha)q + \alpha p).
\end{align}
A weighted Jensen-Shannon divergence is incorporated into their variational objective alongside the standard reverse \gls*{kl} as
\begin{align}\label{eq:UQVAE_objective}
    \phi^{\star} = \underset{\phi}{\argmin}\, J(\phi; \alpha, \vy), \quad 
    J(\phi; \alpha, \vy) = \frac{1}{\alpha}\mathrm{JS}_{\alpha}(q_{\phi}(\vz|\vy)||p(\vz|\vy)) + D_{\KL}(q_{\phi}(\vz|\vy)||p(\vz|\vy)),
\end{align}
where the parameter $\alpha$ allows for a trade-off between data-fit and regularisation, said to help regularise the problem, preventing either extreme low or high values of posterior variance. For expensive forward models, the exact forward model can be replaced by a surrogate decoder $p_{\theta}(\vy|\vz) = \cN(G_{\theta}(\vz), \sigma^2 \mathrm{I})$, $G_{\theta} \coloneqq H^\dagger \circ F_{\theta} \circ \pi_z$ and the encoder and decoder parameters are learned simultaneously.

\subsubsection{Dynamical Latent Spaces}
Embedding dynamical structure into the latent space of a \gls*{vae} has been considered to model time-indexed data $\vy_{1:N} = \{\vy_n\}_{n=1}^{N}$. In \cite{glyn2024varphi}, a probabilistic forward model drives the latent solution, and an auxiliary variable, $\vx_{n}$ is introduced as the \textit{pseudo-observable}, representing the observations of the latent Gaussian state-space model. This yields the likelihoods $p(\vx_n|\vu_n)=\cN(\tilde{H} (\vu_n), \sigma_\vx^2\mathrm{I})$ and $p(\vu_n|\vu_{n-1})=\cN(\Psi^\dagger(\vu_{n-1};\vz), \sigma_{\vu}^2\mathrm{I} ), $
where $\tilde{H}$ is \textit{known} pseudo-observation operator, and $\Psi^\dagger$ is the one-step evolution operator of the latent dynamical system which depends on parameters $\vz$.
The generative model learns to reconstruct data from the pseudo-observable with a probabilistic decoder, $p_{\theta}(\vy_n|\vx_n) = \cN (H_\theta (\vx_{n}), \sigma^2 \mathrm{I})$, where the true mapping is approximated $H^\dagger \approx H_{\theta} \circ \tilde{H}$. The variational posterior is factorised as 
\begin{align}
    q(\vu_{1:N}, \vx_{1:N}, \vz | \vy_{1:N}) \propto p(\vu_{1:N}|\vx_{1:N}) q_{\vartheta}(\vz)\prod_n q_{\phi}(\vx_n|\vy_n),
\end{align}
which uses an amortized encoder $q_{\phi}(\vx_n|\vy_n)$, variational approximation $q_{\vartheta}(\vz)$, and exact posterior $p(\vu_{1:N}|\vx_{1:N})$. We obtain the desired parameters ($\theta^\star, \phi^\star, \vartheta^\star$) by maximising the \gls*{elbo}
\begin{align}
    J(\theta, \phi, \vartheta ; \vy_{1:N}) &= \sum_n \bE_{q_{\phi}(\vx_n|\vy_n)} \left[\log \frac{p_{\theta}(\vy_n|\vx_n)}{q_{\phi}(\vx_n|\vy_n)}\right]\nonumber\\
    &\quad\quad\quad+ \bE_{q_{\phi}(\vx_n|\vy_n)q_{\vartheta}(\vz)}\left[\log p(\vx_{1:N}|\vz)\right] - D_\KL(q_{\vartheta}(\vz)||p(\vz)).
\end{align}
The term $\log p(\vx_{1:N}|\vz)$ is computed using Kalman filtering. Similarly, dynamical latent structure is imposed in \cite{lopez2020variational} by constraining the latent embeddings to non-euclidean manifolds, improving the robustness to noise and improving interpretability of latent dynamics.

\subsubsection{Deep Generative Priors for Regularisation}
When the parameter space is high-dimensional, regularising the inverse problem is essential. Furthermore, if direct observations of the parameters  are available, a possible method of regularisation is through the use of a \gls*{dgp} over the parameter space. By introducing a lower-dimensional auxiliary latent variable $\vw$, a generative model $p_\theta(\vw,\vz)=p_{\theta}(\vz|\vw)p(\vw)$ can be trained to approximately generate samples from the prior $p(\vz)$, where the likelihood is constructed as a probabilistic decoder, e.g. $p(\vz|\vw) = \mathcal{N}(f_{\theta}(\vw), \sigma^2 \mathrm{I})$, with learnable generator function $f_\theta$. Including the \gls*{dgp} in the inverse problem acts as a form of regularisation when optimisation is performed over the low-dimensional $\vw$ rather than the high-dimensional $\vz$. Typically \glspl*{vae} are suitable here~\cite{laloy2017inversion} because of the in-built dimensionality reduction and once trained, the decoder can produce samples from the \gls*{dgp} via $\vz^{(i)} = f_{\theta^\star}(\vw^{(i)}),$ with $\vw^{(i)}\sim p(\vw)$ (here $f_\theta$ need not be invertible). The auxiliary prior can be set arbitrarily, most simply as a standard multivariate Gaussian.

For solving the inverse problem, in \cite{lopez2021deep} {a point-estimate-based} inversion viewpoint is taken, where the optimisation is performed w.r.t. auxiliary variables, which are pushed through the trained generator and then the forward model to obtain the data-misfit loss
\begin{align}
    J(\vw;\vy, \theta^\star) = \Vert G^\dagger\circ f_{\theta^\star}(\vw) - \vy\Vert^2 + \beta (\Vert\vw\Vert - \mu_\chi)^2.
\end{align} 
where the constant $\mu_\chi$ in the regularisation term preferences $\vw$ that lie on a ring centred at the origin. 
The resulting parameter estimate is found by pushing the optimal $\vw^\star = \argmin_{\vw}\; J(\vw;\vy, \theta^\star)$ through the generator, giving $\vz^\star = f_{\theta^\star}(\vw^\star)$.

{
One might consider learning probabilistic priors for inversion through the use of normalizing flows. In~\cite{levy2023variational} the authors trained a normalizing flow to learn a prior in an embedded space -- where the embedding itself is learned with a \gls*{vae} or \gls*{gan}.
}

In \cite{xia2023vi}, a simple \gls*{dgp} is trained for sampling $p(\vz)$, which is included in a Bayesian \gls*{vi} problem where the auxiliary posterior $p(\vw|\vy)$ is approximated by the \gls*{vi} approximation $q_{\phi}(\vw)$. The objective is
\begin{align}
    \phi^\star = \underset{\phi}{\argmin}\;J(\phi;\vy,\theta^\star), \; J(\phi;\vy,\theta^\star) = \bE_{q_{\phi}(\vw)}\left[-\log p(\vy|\vw)\right] + \KL(q_{\phi}(\vw)|p(\vw)),
\end{align}
where the likelihood $p(\vy|\vw)\coloneqq p(\vy|\vz=f_{\theta^\star}(\vw))$ is determined by the forward model, $\vy = G^\dagger \circ f_{\theta^\star}(\vw) + \boldsymbol{\epsilon}$. Posterior samples can then be readily obtained by sampling from this variational posterior and pushing through the generator, $\vz^{(i)} = f_{\theta^\star}(\vw^{(i)}),$ with $\vw^{(i)}\sim q_{\phi^\star}(\vw)$.

\subsection{Residual-Based Learning}\label{subsec:residual_based_learning}
The objective of \Gls*{vi}-based deep surrogate modelling is to predict solutions 
of \glspl*{pde} using deep learning models that output uncertainty about their predictions. Such surrogates are of great use for solving inverse problems as they can replace computationally expensive numerical forward models whilst quantifying the error of their approximations which can be incorporated into inversion schemes~\cite{cleary2021calibrate}.

\subsubsection{Data-Free Inference}

For the work in~\cite{zhu2019physics}, the authors model the PDE solution $\vu$ given a parameter $\vz$ probabilistically through a residual $\vr(u_h, z_h)$ with
\begin{align}\label{eq:res_dist_1}
    p_\beta(\vu|\vz) \propto {\exp\big(-\beta\, \|\vr(\pi_u(\vu), \pi_z(\vz))\|^2_2\big)},
\end{align}
where the exact formulation of the residual $\vr(u_h, z_h)$ can vary, but its purpose remains the same; $\vr= 0$ when $u_h$ satisfies the PDE system for parameters $z_h$.
We then seek the parameters 
\begin{align}
    \phi^\star = \underset{\phi}{\argmin}\;D_\KL(q_\phi(\vu|\vz)p(\vz)||p_\beta(\vu|\vz)p(\vz)),
\end{align}
where $\beta$ controls the intensity of the physics constraint and is selected such that the surrogate model $q_\phi(\vu|\vz)$ provides calibrated uncertainty estimates given a dataset $\mathcal{D}=\{\vu^{(n)}, \vz^{(n)}\}_{n=1}^N$ of solution-parameter pairs. In their work, the authors make use of a normalizing flow to model the forward problem $q_\phi(\vu|\vz)$.  
This variational construction 
{learns a probabilistic forward model}.

In~\cite{vadeboncoeur2023fully, vadeboncoeur2023random} different variational frameworks are proposed which allow for the learning of both forward and inverse {probabilitic maps}.
The construction is posed through a parametrised probabilistic model $p_\theta(\hat{\vr}, \vu,\vz) = p(\hat{\vr}|\vu,\vz)p_\theta(\vz|\vu)p(\vu)$ and a variational approximation $q_\phi(\vu,\vz)=q_\phi(\vu|\vz)q(\vz)$.
Here, $\hat{\vr}$ represents a zero-valued \textit{virtual observable}~\cite{rixner2021probabilistic} posed as
\begin{align}
    \hat{\vr} = \vr(\pi_u(\vu), \pi_z(\vz)) + \boldsymbol{\epsilon}_r, \quad {\boldsymbol{\epsilon}}_r\sim\cN(0, \sigma_r^2\mathrm{I}).
\end{align}
We note that other virtual noise models may be considered, leading to different residual likelihoods~\cite{chatzopoulos2024physics}.
The factorization of the joint variational approximation $q_\phi(\vu, \vz)$ and the model $p_\theta(\vu,\vz|\hat{\vr})$ is chosen such that
\begin{align}
\label{eq:ELBO_without_data}
         \phi^\star, \theta^\star = \underset{\phi,\theta}{\argmax}\; J(\phi,\theta),\quad J(\phi,\theta) = \bE_{q_{\phi}(\vu | \vz) p(\vz)} \log \frac{p(\hat{\vr} = 0|\vu, \vz) p_{\theta}(\vz | \vu) p(\vu)}{q_{\phi}(\vu | \vz) p(\vz)},
\end{align}
learns mapping for forward UQ ($ q_{\phi}(\vu | \vz)$) and inversion ($p_{\theta}(\vz | \vu)$). It is a lower bound on the log marginal probability of $\hat{\vr}$. In the same spirit as~\eqref{eq:res_dist_1} (with $\beta = 1/2\sigma_r^2$), the distribution over the residual is posed as $p(\hat{\vr}=0|\vu, \vz)\propto \exp(-\frac{1}{2\sigma_r^2}\|\vr(\pi_u(\vu), \pi_z(\vz))\|^2_2)$. These frameworks construct variational uncertainty quantifying surrogates in the data-free regime. 

\subsubsection{Small-Data Regime}

In some settings, one may have access to small datasets alongside knowledge of the form of the underlying physics. Methods for constructing probabilistic forward surrogates may pose their likelihood as a product measure between a virtually observed residual $\hat{\vr}$ and data $\vy$ as in~\cite{kaltenbach2020incorporating}. Using this approach, one can combine (possibly high-fidelity) data with fast to evaluate physics residuals in the likelihood
\begin{align}
    p(\hat{\vr}, \vy|\vu, \vz) = p(\hat{\vr}=0|\vu, \vz)p(\vy|\vu, \vz),
\end{align}
where the balance between data and physics residual is given by the estimated variance of the data noise and chosen virtual observational noise of the residual. A Bayes \gls*{vi} objective can be written using ~\eqref{eq:J_Bayes_VI} to obtain an approximate posterior over the solution $\vu$ and parameters $\vz$ as
\begin{align}
    \phi^\star = \underset{\phi}{\argmin}\; J(\phi),\quad J(\phi) = D_\KL(q_\phi(\vu,\vz)||p(\vu, \vz|\vy, \hat{\vr})).
\end{align}
Here $q_\phi(\vu,\vz)$ is factorized independently as $ q_\phi(\vu)q_\phi(\vz)$ -- called the mean field approximation~\cite{parisi1988statistical} -- and the dependence between the parameter and solution to the PDE is captured in the likelihood through the virtual observable $\hat{\vr}=0$. Similar in objective is~\cite{tait2020variational}, where a joint variational approximation $q_\phi(\vu, \vz)$ is used to approximate the Bayesian posterior $p(\vu, \vz|\vy)$, factorising $q_\phi(\vu, \vz) = q_\phi(\vu|\vz) q_\phi(\vz)$ where the likelihood $q_\phi(\vu|\vz)$ = $\cN(\vu;F_\phi(\vz), \epsilon^2 C(\vz))$ captures the forward map. Furthermore,  \cite{tait2020variational} uses the information from the physics problem through the stiffness matrix to inform the covariance $C(\vz)$. The parameter $\epsilon$ controls the strength of the physics constraint in the likelihood, and in the limit $\epsilon\rightarrow 0$, the following problem is recovered 
\begin{subequations}
\begin{align}
    \theta^\star, \phi^\star &= \underset{\theta, \phi}{\argmin}\; \bE_{q_\phi(\vz)}\left[ -\log p(\vy|\vu=F_\theta(\vz))\right] + D_\KL(q_\phi(\vz)||p(\vz)),\\
    &\text{s.t. }\; \|\vr(\pi_u(F_\theta(\vz)), \pi_z(\vz))\|^2_2 = 0.
\end{align}
\end{subequations}
Notice in this interpretation, the learning of $F_\theta(\vz)$ is part of the \textit{probabilistic model} not the \textit{variational approximation}, hence changing $F_\phi$ for $F_\theta$. This constrained optimization view is in effect similar to having access to the forward model $F^\dagger$. In~\cite{wang2022variational} a deterministic forward surrogate $F_\theta\approx F^\dagger$ is learned by minimizing $\|(F_\theta - F^\dagger)\circ\pi_z(\vz)\|^2_2$  in conjunction with a normalizing flow which probabilistically solves the inverse problem.
We note that for many of these inversion methods, amortization could be used to learn a mapping to the posteriors given data from varying physical systems.
Relevant to the aforementioned methods, the work in~\cite{grigo2019physics} uses \gls*{vi} to synthesise information for coarse-grained models in the small-data regime. This model is also used to learn efficient latent representations of structured high-dimensional feature spaces, arising in problems in porous-media~\cite{dasgupta2024dimension}. Further methods propose \gls*{vi} surrogate models in the small data-regime for related applications~\cite{rixner2022self}.

{        Methods for handling stochastic \glspl*{pde} have also been developed to solve forward and inverse problems when the solution, parameters, and source terms are described by random fields. These fields may only be sparsely observed over a number of sensor locations. The variational autoencoder approaches in~\cite{shin2023physics,zhong2023pi} encode observations to auxiliary random variables, which capture the stochastic behaviour of the \glspl*{pde}, with physics-informed losses constructed from PDE residual terms. Aside from \glspl*{vae}, other \gls*{vi} variants include Physics-Informed generative adversarial networks (PI-GAN)~\cite{yang2020physics}, and normalizing field flows (NFF)~\cite{guo2022normalizing}, which uses physics-informed flows and is agnostic to sensor/observation location.}

\section{Discussion}

This paper introduces the core concepts necessary for constructing \gls*{vi} schemes for solving physics-based forward and inverse problems. Furthermore, we review the literature that employs \gls*{vi} and deep learning in the context of physics, presenting the contributions under a unified notation. Our approach is intended to help readers better understand the similarities, differences, and nuances among the various methodologies proposed in the field.
A few limitations are to be kept in mind when applying and developing some of the mentioned works.
As highlighted in~\cite{zhu2019physics}, care must be taken in assessing the {accuracy} of uncertainty quantification with \gls*{vi}, which remains an open practical~\cite{povala2022variational} and theoretical challenge~\cite{wang2019frequentist}. In applications, one should also assess the computational advantage of \textit{training} any surrogate model versus directly making use of classical numerical schemes~\cite{de2022cost}.  
Software libraries are being developed  to aid practitioners in the implementation of these schemes eg.~\cite{zou2024neuraluq}.
Furthermore, the use of the KL divergence may not always be well-posed, particularly when dealing with functional objects such as in physics applications~\cite{bunker2024autoencoders}.
As such, beyond the Bayes formulation of \gls*{vi}, promising areas of research consider other divergences~\cite{knoblauch2019generalized} such those based on the Wasserstein~\cite{ambrogioni2018wasserstein, yao2022mean} and Sliced Wasserstein metrics~\cite{akyildiz2024efficient, yi2023sliced} or Maximum Mean Discrepancy~\cite{cherief2020mmd, zhong2023pi} as these do not have the same conditions on absolute continuity and are readily computable from random samples. Finally, many promising developments in solving physics-based inverse problems through deep learning and possibly variational inference focus on learning better priors~\cite{patel2022solution, patel2021gan, akyildiz2024efficient, meng2022learning} { along with important earlier works in Earth sciences~\cite{laloy2017inversion, laloy2018training}.} 


\noindent\textbf{Funding:}{AGD is supported by Splunk Inc. [G106483] PhD scholarship funding.
AV is supported through the EPSRC ROSEHIPS grant [EP/W005816/1].
MG is supported by a Royal Academy of Engineering Research Chair and EPSRC grants [EP/X037770/1, EP/Y028805/1,
EP/W005816/1, EP/V056522/1, EP/V056441/1, EP/T000414/1, EP/R034710/1].}

\bibliography{bibliography}

\end{document}